\documentclass[letterpaper, 10 pt, conference]{ieeeconf}  

\IEEEoverridecommandlockouts                              

\usepackage{graphics} 
\usepackage{epsfig} 
\usepackage{mathptmx} 
\usepackage{times} 
\usepackage{amsmath} 
\usepackage{amssymb}  
\usepackage{booktabs}
\usepackage{hyperref}
\usepackage{mathtools}
\renewcommand{\autoref}[1]{Fig.~\ref{#1}} 

\usepackage{soul} 
\usepackage{xcolor} 

\definecolor{dred}{rgb}{0.8,0,0}
\newcommand{\alert}[1]{\textcolor{dred}{\hl{\textbf{#1}}}}
 \renewcommand{\alert}[1]{}\renewcommand{\hl}[1]{} 

\title{\LARGE \bf
Safe Execution of RL Policies via Acceleration-based CBF-QP Constraint Enforcement for Real-World Robotic Deployments
}

\author{Bastien Muraccioli$^{*}$$^{1}$, Alice Cariou$^{*}$$^{1}$, Pierre-Alexandre Leziart$^{1}$, Mathieu Celerier$^{1}$, Arnaud Demont$^{1}$,\\ Gentiane Venture$^{1,2}$, and Mehdi Benallegue$^{1}$
\thanks{$^{*}$B. Muraccioli, A. Cariou share first authorship.}%
\thanks{$^{1}$All authors are with the CNRS–AIST Joint Robotics Laboratory (JRL), IRL3218, AIST, Tsukuba, Japan, {\tt\small\{bastien.muraccioli, alice.cariou, pa.leziart, mathieu.celerier, arnaud.demont, mehdi.benallegue\}@aist.go.jp}}%
\thanks{$^{2}$G. Venture is also with the Department of Mechanical Engineering, The University of Tokyo, Japan, {\tt\small venture@g.ecc.u-tokyo.ac.jp}}}%

\begin{document}

\maketitle
\thispagestyle{empty}
\pagestyle{empty}

\begin{abstract}
Reinforcement Learning (RL) has demonstrated remarkable capabilities for solving complex robotic control problems, but its lack of safety guarantees severely limits deployment on hardware. In particular, as legged robots and manipulators often operate near safety-critical boundaries, out-of-distribution states can lead to failure upon deployment. To address this, we introduce Acc-CBF-QP, an acceleration-based Quadratic Program (QP) safety filter using Control Barrier Functions (CBFs) that constrains any RL policy onto a safe set at runtime without modifying training. The method applies to unconstrained and Safe-RL policies, and enforces joint position, velocity, torque, and collision constraints within a unified optimization framework. A key contribution is the formulation of RL+QP tasks that regulate deviation from the RL command when constraints would otherwise be violated. We introduce a \emph{TorqueTask}, minimizing torque deviation, and a \emph{Forward Dynamics Task}, minimizing induced acceleration deviation, thus providing principled control over safety-performance trade-offs. Experiments on a 7-DoF Kinova Gen3 manipulator and a 19-DoF Unitree H1 humanoid, both in simulation and on hardware, highlight substantial reductions in constraint violations. On the real H1 hardware, a Safe-RL policy alone yielded 10.04 violations/s, which were reduced by 92\% to 0.80 violations/s when augmented with Acc-CBF-QP. On the Kinova Gen3, Acc-CBF-QP fully eliminated violations. Nominal task performance of the RL objective is preserved in violation-free regimes. Under aggressive velocity commands on H1, Acc-CBF-QP improves execution by preventing constraint-induced shutdowns, yielding longer survival times. The full pipeline is open-source.
\end{abstract}

\section{Introduction}

Reinforcement Learning (RL) has emerged as a powerful tool for robotics, enabling the autonomous acquisition of complex skills such as agile locomotion and dexterous manipulation through trial-and-error interactions. Despite promising results, a persistent obstacle to deploying RL policies on hardware is the lack of \emph{safety guarantees}. In this work, we define safe execution of policies as their ability to respect explicit constraints. Such constraints can be imposed for feasibility, for preventing hardware damage and adhering to the robot manufacturer’s specifications, and for human and environmental safety.

\begin{figure}[t]
    \centering
    \includegraphics[width=0.40\textwidth]{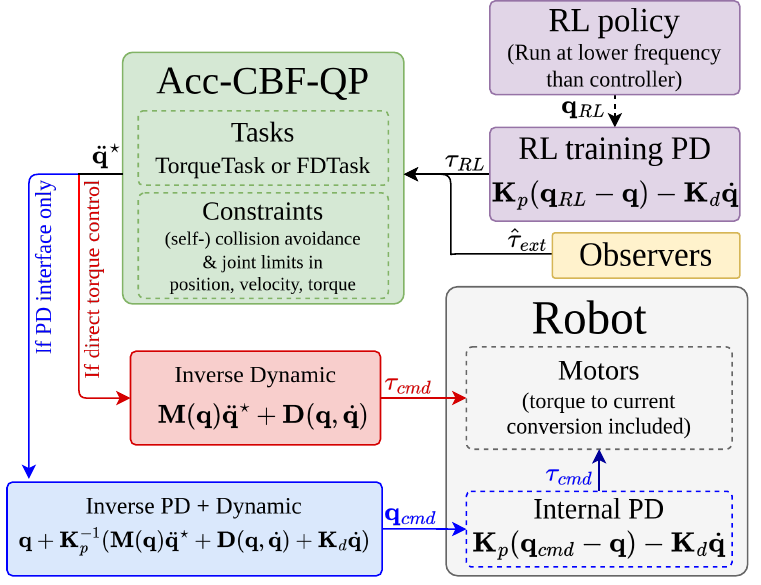}
    \caption{Controller pipeline. The RL policy (purple) is filtered in real time by an Acc-CBF-QP (green) that enforces safety constraints using external disturbance estimates from observers (yellow). The resulting feasible commands are applied in torque control (red) or with a position reference for a Proportional-Derivative (PD) controller (blue) to the robot (gray).}
    \label{fig:controller-pipeline}
\end{figure}

Safe RL methods embed constraints during training~\cite{10.24963/ijcai.2023/763}, via constrained Markov Decision Processes~\cite{altman2021constrained, hung2025efficient}, Control Barrier Functions~\cite{8796030, 9928337}, or probabilistic safety filters~\cite{10801338}. While effective in simulation, these rely on statistical guarantees that degrade under distributional shift, require retraining when constraints change, and often fail to hold on Out-of-Distribution (OoD) states during deployment.

Furthermore, enforcing strict safety constraints during training can be overly conservative, restricting exploration, slowing convergence~\cite{ai6030046}, and limiting adaptability when new safety requirements emerge at deployment, such as collision avoidance based on online perception.

In contrast, the robotics community has long relied on optimization-based controllers that enforce constraints for feasibility or safety in real time. In particular, Quadratic Program (QP) control in acceleration space is well established for whole-body control, enabling prioritized task execution under joint, torque, and contact constraints~\cite{7803308}, and safety for Human-Robot Interaction (HRI)~\cite{muraccioli2025rss}. However, such approaches require model-based tasks and do not directly integrate with learned policies, limiting their adaptability.

\textbf{Our approach bridges these two paradigms.} Building on the acceleration-based QP-CBF framework of~\cite{muraccioli2025rss}, we extend it to accept RL-policy torques as the nominal reference and validate it as a post-processing safety filter around any pre-trained RL policy. \autoref{fig:controller-pipeline} illustrates the overall controller pipeline. At each control step, the RL command is passed through the QP, which enforces constraint boundaries formulated as CBFs. The resulting optimization provides forward-invariance guarantees under feasibility and bounded-modeling-error assumptions, formalized in Sec.~\ref{subsection:formal}. This formulation enables runtime constraint enforcement without requiring additional policy retraining. This is a critical advantage, as it avoids restricting exploration during training and permits dynamic modification of safety constraints at deployment.
The resulting constrained behavior can be optimized for different criteria, such as minimizing torque error or acceleration error, depending on the application, as detailed in Section~\ref{subsection:rl-qp-tasks}.

\textbf{Contributions.} The main contributions of this work are:
\begin{itemize}
\item An open-source framework\footnote{https://safe-rl-qp.github.io/} for enforcing safety on robotics RL policies through an acceleration-based QP with CBF-based constraints (Acc-CBF-QP), featuring two task formulations (TorqueTask and FDTask) that control how the QP minimizes deviation from the RL objective when constraints are active. The pipeline integrates state estimation and disturbance observers to recover external forces and mitigate modeling errors, essential for consistent CBF enforcement on torque-controlled and floating-base systems.
\item Experimental validation on two distinct robot platforms: the Kinova Gen3 manipulator and the Unitree H1 humanoid, across both simulation and real hardware. Results demonstrate substantial reductions in constraint violations and hardware-damaging failures, with quantified trade-offs between safety enforcement and task performance. On the humanoid, we compare against a CaT-based Safe RL baseline~\cite{chane2024cat}.
\end{itemize}
This work establishes a modular interface between learned control and acceleration-level safety filtering, constituting a practical step toward the reliable real-world deployment of learned robotic behaviors.

\section{Related Works}

Safety in robot learning spans both \emph{training-time approaches}, which attempt to embed safety into the policy learning process, and \emph{execution-time filters}, which modify unsafe commands during deployment. Our work contributes to the latter category, building on optimization-based safety layers.

\subsection{Safe Reinforcement Learning}
Safe RL methods typically formulate safety as constraints in a CMDP and enforce them during learning, such as Constrained Policy Optimization (CPO), which applies a trust-region constrained update that provides per-iteration safety guarantees~\cite{achiam2017constrained}. While these approaches incorporate constraints directly into policy optimization by adjusting their update rules to discourage violations, others include action-constrained methods that ensure feasibility through acceptance–rejection sampling and augmented Markov Decision Processes~\cite{hung2025efficient}, and optimization-based RL that leverages QP/MPC-style policy parameterizations to obtain formal guarantees during training~\cite{zanonReinforcementLearningGuarantees2025,TORRE202442}. Recent work also explores learning on constraint manifolds for improved safety in robotic systems~\cite{liuSafeReinforcementLearning2024}. 
Among training-time approaches, the Constraints as Terminations (CaT) framework~\cite{chane2024cat} enforces safety by introducing stochastic termination signals upon constraint violation, encouraging constraint satisfaction during learning. However, like other training-time methods, it provides distributional guarantees and does not ensure consistent per-state constraint satisfaction at inference. We adopt CaT as a Safe RL baseline to contrast training-time safety shaping with runtime formal filtering.

While these methods provide guarantees in control-theoretic settings, they remain challenging to apply concretely in robotics due to unmodeled dynamics and their reliance on highly accurate system models. Our approach also relies on a system model, but to a lesser extent: model inaccuracies are mitigated through a disturbance observer (DOB), which estimates and compensates for unknown loads~\cite{motioncontrol}. This allows the safety filter to remain effective even under imperfect modeling. 

\subsection{Safety Filters and Runtime Enforcement}
To ensure constraint satisfaction during deployment, safety filters act as post-processing layers around RL policies. Notable examples include predictive safety filters~\cite{wabersich2021predictivesafetyfilterlearningbased} and CBF-CLF-QP-based safety layers~\cite{cbf-cqf, shield2025}. These methods provide varying levels of safety (hard, soft, or probabilistic). Yet, they were either tested in simulation only, or not tested with physical constraints such as joint limits or collisions on the learned policy output.

\subsection{Positioning of This Work}
Acc-CBF-QP acts as a runtime safety layer that composes a learned torque policy with acceleration-level CBF constraints through structured optimization. At each control step, the QP minimizes deviation from the RL objective while satisfying all CBF constraints. This mechanism can be interpreted as a state-dependent projection of the RL torque onto a CBF-defined admissible set (formalized in Sec.~\ref{subsection:formal}). Unlike training-time Safe RL, which promotes constraint satisfaction in expectation, Acc-CBF-QP enforces it consistently under QP feasibility, without retraining.

\section{Method}

\subsection{Robot Dynamics and Notation} 
We consider an $n$-DoF multi-body rigid robot, whose configuration in space can be fully described by its generalized coordinates denoted $\mathbf{q} \in \mathbb{R}^n$. 
Let $n_{\text{a}}$ denote the number of actuated joints, with actuated coordinates $\mathbf{q}_{\text{a}}\in\mathbb{R}^{n_{\text{a}}}$ and controlled with a torque $\boldsymbol{\tau}_{\text{a}}\in\mathbb{R}^{n_{\text{a}}}$.
The dynamics under external generalized forces/torques $\boldsymbol{\tau}_{\text{ext}}\in\mathbb{R}^n$ are
\begin{equation}
 \boldsymbol{\tau} = \mathbf{M}( \mathbf{q}) \ddot{ \mathbf{q}} + \mathbf{C}(\mathbf{q}, \dot{\mathbf{q}})\dot{\mathbf{q}} + \mathbf{g}(\mathbf{q}) - \boldsymbol{\tau}_{\text{ext}}, \label{eq:dynamicmodel} 
\end{equation}
and 
\begin{equation}
    \boldsymbol{\tau} = \mathbf{S}^{\top} \boldsymbol{\tau}_{\text{a}}.
\end{equation}
$\mathbf{M}(\mathbf{q}) \in \mathbb{R}^{n \times n}$ is the robot's inertia matrix, $\mathbf{C}(\mathbf{q}, \dot{\mathbf{q}}) \in \mathbb{R}^{n \times n}$ is the Coriolis and centrifugal effects, and $\mathbf{g}(\mathbf{q}) \in \mathbb{R}^n$ is the gravitational torque vector.
$\mathbf{S} \in \mathbb{R}^{n_{\text{a}}\times n}$ is the \emph{actuated joint selection matrix}, such that $\dot {\mathbf{q}}_{\text{a}} = \mathbf{S} \dot{\mathbf{q}}$~\cite{sentis2005controlFreeFloatingBase, bouyarmane2018quadraticProgrammingMultiRobot}. 

For a fixed-base, fully actuated robot, the generalized coordinates coincide with the joint coordinates,
$\mathbf{q}=\mathbf{q}_{\text{a}}$ and $n=n_{\text{a}}$, hence $\mathbf{S}$ is simply the $n\times n$ identity matrix.

For a floating-base robot with fully actuated joints, which moves freely in space, the generalized coordinates also include the pose of the floating base in the world $\mathbf{q}_{\text{fb}}\in\mathrm{SE}(3)$, represented with any parameterization of size $n_\text{fb}$:
\begin{equation} 
    \mathbf{q} = 
    \begin{bmatrix}\mathbf{q}_{\text{fb}} \\ \mathbf{q}_{\text{a}}\end{bmatrix} 
    \in \mathbb{R}^{n_{\text{fb}}+n_{\text{a}}} .
\end{equation}
The external generalized forces/torques $\boldsymbol{\tau}_{\text{ext}}$ now include the external wrench applied on the floating base. For simplicity, following~\cite{bouyarmane2018quadraticProgrammingMultiRobot}, we assume $\mathbf{q}$, $\dot{\mathbf{q}}$, $\ddot{\mathbf{q}}$ all have dimension $n$, treating the floating-base velocity and acceleration as their SE(3) counterparts by standard abuse of notation.
  
\subsection{Acceleration-Based CBF-QP Safety Layer}\label{subsection:QP}
Formulating the control problem as a QP in accelerations enables the simultaneous satisfaction of dynamic and kinematic constraints~\cite{torque-control-QP}. Many robotic constraints are second-order in nature (e.g., acceleration limits, torque limits, or contact force constraints). By choosing the accelerations as the decision variables, these constraints become linear or convex, allowing efficient QP solvers to find feasible solutions. This motivates the use of an acceleration-based QP to enforce model-based constraints while executing an RL policy.

\subsubsection{Acceleration-based QP formulation}
At each control step, the QP computes the generalized accelerations $\ddot{\mathbf{q}}^\star \in \mathbb{R}^{n}$ that satisfy all constraints while minimizing a quadratic cost written as a weighted sum of task errors:
\begin{align}\label{eq:QP}
\ddot{\mathbf{q}}^\star 
&= \arg\min_{\ddot{\mathbf{q}}} \;
   \sum_k\|\mathbf{W}_k(\mathbf{A}_k\ddot{\mathbf{q}} - \mathbf{b}_k)\|^2 \\
\text{s.t.}\quad & \mathbf{A}_c \ddot{\mathbf{q}} \leq \mathbf{b}_c, \nonumber
\end{align}
where $(\mathbf{A}_k, \mathbf{b}_k)$ encode the linearized tasks, $(\mathbf{A}_c, \mathbf{b}_c)$ encode the linearized constraints, and $\mathbf{W}_k \succ 0$ acts as a weight matrix on the corresponding task. 

This is standard in QP-based control. Although the formulation supports multiple simultaneous tasks, we deliberately evaluate one task $(\mathbf{A},\mathbf{b})$ at a time to isolate its contribution without confounding effects from task prioritization or $\mathbf{W}_k$ tuning; combining tasks (e.g., locomotion with manipulation) is left for future work.

\subsubsection{Constraint set}
Safety is defined as the forward invariance of the closed set
\begin{equation}
\mathcal C =
\bigcap_{i=1}^{N_c}
\{(\mathbf q,\dot{\mathbf q}) \mid h_i(\mathbf q,\dot{\mathbf q}) \ge 0 \},
\end{equation}
where $N_c$ is the number of safety constraints and each barrier function $h_i$ is continuously differentiable. For vector-valued bounds, inequalities are understood componentwise. 
Assuming perfect acceleration tracking, forward invariance is enforced via Control Barrier Function (CBF) conditions~\cite{expo-cbf,cbf-joint}. 
Since the QP decision variable is $\ddot{\mathbf q}$, all barrier conditions are expressed as affine inequalities in $\ddot{\mathbf q}$ and stacked into $(\mathbf A_c,\mathbf b_c)$.

\paragraph{Torque limits}
Actuator limits are enforced at the torque level as
\begin{equation}\label{eq:tau-constraint}
\boldsymbol{\tau}_{\text a}^- \le \boldsymbol{\tau}_{\text{QP}} \le \boldsymbol{\tau}_{\text a}^+,
\end{equation}
where $\boldsymbol{\tau}_{\text{QP}}$ is obtained from $\ddot{\mathbf q}^\star$ via inverse dynamics (Eq.~\eqref{eq:tau-QP}). For floating-base robots, torques associated with the base are constrained to zero.

\paragraph{Joint velocity and position limits}
Consider an upper velocity bound $\dot{\mathbf q}_+$, such that $\dot{\mathbf q}\le \dot{\mathbf q}_+$. Define the barrier function 
\begin{equation}
h_{vel}(\dot{\mathbf q}) = \dot{\mathbf q}_+ - \dot{\mathbf q}.
\end{equation} 

Since $h_{vel}$ has relative degree one with respect to $\ddot{\mathbf q}$, the standard CBF condition $\dot h_{vel} + \alpha(h_{vel}) \ge 0$ is imposed, where $\alpha$ is a class-$\mathcal K$ function. Choosing the linear function $\alpha(h_{vel})=\lambda h_{vel}$ with $\lambda>0$, as done by~\cite{muraccioli2025rss}, yields
\begin{equation}\label{eq:vel-cst}
\ddot{\mathbf q} \le -\lambda(\dot{\mathbf q}-\dot{\mathbf q}_+).
\end{equation}
Lower velocity bounds are treated analogously.

For an upper position limit $\mathbf q_+$, such that $\mathbf q \le \mathbf q_+$, define \begin{equation}
h_{pos}(\mathbf q)=\mathbf q_+ - \mathbf q.
\end{equation}
Since $h_{pos}$ has relative degree two with respect to the control input, we employ an exponential CBF (ECBF)~\cite{expo-cbf}, enforcing \begin{equation}
\ddot h_{pos} + \alpha_1 \dot h_{pos} + \alpha_2 h_{pos} \ge 0,
\end{equation} 
with $\alpha_1,\alpha_2>0$. Using $\dot h_{pos}=-\dot{\mathbf q}$ and $\ddot h_{pos}=-\ddot{\mathbf q}$, this yields 
\begin{equation}
\ddot{\mathbf q}
\le
-(\alpha_1+\alpha_2)\dot{\mathbf q}
+
\alpha_1\alpha_2(\mathbf q_+ - \mathbf q).
\end{equation} 
Following~\cite{muraccioli2025rss}, we parametrize $\alpha_1 + \alpha_2 = \lambda$ and $\alpha_1\alpha_2 = \lambda^2/(4\zeta^2)$ with $\zeta \ge 1$, yielding
\begin{equation}\label{eq:joint-cst}
\ddot{\mathbf q} \le -\lambda \dot{\mathbf q}+\frac{\lambda^2}{4\zeta^2}(\mathbf q_+ - \mathbf q),
\end{equation}
where $\zeta=1$ gives critical damping and $\zeta>1$ overdamped behavior. Sharing $\lambda$ across velocity and position limits enforces consistent convergence rates and simplifies tuning. Lower bounds are handled analogously.

\paragraph{Self-collision avoidance}
For each monitored pair of bodies, let $e(\mathbf q)$ denote their minimum distance and define the barrier function
\begin{equation}
h_{col}(e) = e - d_s,
\end{equation}
where $d_s > 0$ is a prescribed safety margin. 
Assuming continuous motion of the closest points, $e(\mathbf q)$ is differentiable locally everywhere and satisfies
\begin{equation}
\dot e = \mathbf J_e(\mathbf q)\dot{\mathbf q}, 
\qquad
\ddot e = \mathbf J_e(\mathbf q)\ddot{\mathbf q}
+ \dot{\mathbf J}_e(\mathbf q,\dot{\mathbf q})\dot{\mathbf q},
\end{equation}
where $\mathbf J_e(\mathbf q)=\frac{\partial e}{\partial \mathbf q}$ denotes the distance Jacobian, obtained by projecting the geometric Jacobians of the closest points onto the inter-body normal direction, as done in ~\cite{qp-constraint, vaillant2016multi}.

As $h_{col}$ has relative degree two with respect to $\ddot{\mathbf q}$, we employ the same exponential CBF construction as for joint position limits, yielding
\begin{equation}
\mathbf J_e(\mathbf q)\ddot{\mathbf q}
\ge
-\dot{\mathbf J}_e(\mathbf q,\dot{\mathbf q})\dot{\mathbf q}
-\lambda \dot e
-\frac{\lambda^2}{4\zeta^2}(e-d_s),
\end{equation}
which is affine in $\ddot{\mathbf q}$. 
Collision-avoidance gains $(\lambda, \zeta)$ are tuned independently from joint-limit gains.

\medskip

All actuator and barrier constraints are finally stacked into $(\mathbf A_c,\mathbf b_c)$ and enforced within the QP at each control step. Feasibility of the QP is therefore a necessary condition for formal safety guarantees. CBF gains, safety margins, and observer gains were tuned empirically per platform, and representative values are given in the open-source implementation\footnotemark[1].

\subsubsection{External disturbance estimation}\label{para:external-force}
To accurately predict the dynamics of the robot  (Eq.~\eqref{eq:dynamicmodel}), we estimate the external disturbance $\hat{\boldsymbol{\tau}}_\text{ext}$ with a generalized-momentum observer (GMO)~\cite{residual-arm}. This estimate behaves as a first-order filtered signal of external torques (including contacts), torque tracking errors, and other dynamic modeling errors, combined into a single generalized torque vector. When available, force–torque sensors can improve high-frequency estimation~\cite{ralcelerier}.

Since GMO requires generalized velocity signals, which are not all measured on floating-base robots, we implement a base velocity estimator based on the Tilt Observer~\cite{tilt}. This complementary filter estimates the IMU tilt and linear velocity with proven global asymptotic convergence, aligning with our safety-oriented objectives.
The Tilt Observer requires gyrometer and accelerometer, together with a rough linear velocity, to prevent drift. As suggested in~\cite{tilt}, for legged robots, this signal can be obtained from an \emph{anchor point}, defined as a point attached to the robot with zero velocity in the world frame. In practice, assuming no contact slippage, the anchor point is computed from the average contact position, weighted to favor contacts better satisfying Coulomb friction assumptions. 
In our experiment, ground reaction force was not readily available, since the Unitree H1 is not equipped with force sensors. Thus, we used instead the estimated external torque at each knee, assuming it to be proportional to the vertical reaction force.
Note that even if this scheme makes the Tilt Observer and GMO mutually dependent, it has the merit of providing an estimate $\hat{\boldsymbol{\tau}}_\text{ext}$ of external disturbances and modeling errors on a floating-base robot using only IMU and joint encoder signals. 
As a first-order filter, the GMO trades estimation delay against noise rejection; the resulting phase lag during foot-ground impact likely contributes to the residual $\mathbf e_\tau$ (Sec.~\ref{subsection:formal}) and the joint-velocity violations of Table~\ref{tab:constraint_violations}.

Using this estimate, we can finally compute:

\begin{equation}\label{eq:tau-QP}
    \boldsymbol{\tau}_\text{QP} =   
    \mathbf{M}(\mathbf{q}) \ddot{\mathbf{q}}^\star 
    + \mathbf{C}(\mathbf{q}, \dot{\mathbf{q}})\dot{\mathbf{q}} + \mathbf{g}(\mathbf{q}) - \hat{\boldsymbol{\tau}}_{\text{ext}},
\end{equation}
which is the expression used for torque-limit constraints (Eq.~\eqref{eq:tau-constraint}), and as the torque command.

\subsubsection{Robustness analysis}\label{subsection:formal}
Combining Eq.~\eqref{eq:tau-QP} with the true dynamics~\eqref{eq:dynamicmodel} gives
\begin{equation}\label{eq:qddot-real}
\ddot{\mathbf q}_{real}
=
\ddot{\mathbf q}^\star
+
\mathbf M(\mathbf q)^{-1}\mathbf e_\tau,
\qquad
\mathbf e_\tau
:=
\boldsymbol\tau_{ext}
-
\hat{\boldsymbol\tau}_{ext}.
\end{equation}
If, for some known disturbance bound $\bar\delta\ge0$,
\begin{equation}\label{eq:bounded-error}
\|\mathbf M(\mathbf q)^{-1}\mathbf e_\tau\|_\infty \le \bar\delta,
\end{equation}
propagating this disturbance through the relative-degree-one CBF condition of the joint velocity constraint~\eqref{eq:vel-cst} yields, via the comparison lemma,
\begin{equation}\label{eq:issf}
h_{vel}(t)\ge
h_{vel}(0)e^{-\lambda t}
-
(\bar\delta/\lambda)(1-e^{-\lambda t}).
\end{equation}
This provides an \emph{input-to-state safety}~\cite{issf} guarantee: the velocity-constrained trajectory remains within an $O(\bar\delta/\lambda)$ neighborhood of the safe set, rather than strictly inside it. An analogous but distinct bound, with a margin scaling as $\bar\delta/(\alpha_1\alpha_2)$, holds for the relative-degree-two position and collision barriers. This is consistent with the nonzero hardware violations reported in Table~\ref{tab:constraint_violations}, compared with zero violations in simulation where $\mathbf e_\tau\approx0$. If the QP becomes infeasible, neither this robustness guarantee nor exact forward invariance applies, and safety instead relies on the fallback mechanisms described in Sec.~\ref{sec:limitations}.

\subsection{RL Control Law Principle}\label{subsection:rl-law}

We consider the classical setup in which a reinforcement learning (RL) policy generates joint-space commands for a robot. 
Most RL policies output an action vector $\mathbf{a} \in \mathbb{R}^{n_{\text{a}}}$, which is typically interpreted as a desired joint displacement relative to a reference configuration $\mathbf{q}_0\in \mathbb{R}^{n_{\text{a}}}$ used during training:
\begin{equation}\label{eq:qRL}
   \mathbf{q}_{\text{RL}} = \mathbf{q}_0 + \mathbf{a},
\end{equation}
where $\mathbf{q}_{\text{RL}}\in \mathbb{R}^{n_{\text{a}}}$ denotes the commanded joint position. 

On real robots, position commands are handled by low-level motor controllers, which convert the desired positions into torque (and ultimately current) commands to drive the actuators.
These inner loops typically run at a higher frequency than the outer policy.
In simulation, this behavior is commonly emulated using a joint-level Proportional-Derivative (PD) controller:
\begin{equation}\label{eq:tauRL}
   \boldsymbol{\tau}_{\text{RL}} = \mathbf{K}_p \bigl( \mathbf{q}_{\text{RL}} - \mathbf{q}_{\text{a}} \bigr) - \mathbf{K}_d \dot{\mathbf{q}}_{\text{a}}.
\end{equation}
where $\mathbf{K}_p, \mathbf{K}_d \in \mathbb{R}^{n_{\text{a}}\times n_{\text{a}}}$ are diagonal gain matrices. 
As a result, the RL policy learns in closed-loop with this controller: the policy effectively observes the combined effect of its commands and the PD dynamics, and the two cannot be treated independently. Consequently, the torque actually realized on the robot, whether in simulation or on hardware, corresponds to $\boldsymbol{\tau}_{\text{RL}}$.

\subsection{Integration of RL with QP Tasks}\label{subsection:rl-qp-tasks}

As established in the previous subsection, the effective output of the RL policy is not the commanded joint position $\mathbf{q}_{\text{RL}}$ but $\boldsymbol{\tau}_{\text{RL}}$.
Therefore, to reproduce the RL behavior within the QP framework, we must formulate linear acceleration tasks that depend on torque. 
In this work, we consider two alternatives: a \emph{Torque task} (TorqueTask), which minimizes the error between the desired torque and the commanded one, and a \emph{Forward Dynamics Task} (FDTask), which minimizes the difference between the QP-predicted acceleration and the one implied through forward dynamics from the desired torque. Other choices (e.g., force or momentum tasks) could be integrated in the same framework, but are left for future work. The derivation in Section~\ref{subsection:rl-law} assumes a joint-space PD law, as this is a common formulation in RL training frameworks. However, the proposed integration only requires that the RL policy induce a joint-torque vector. For example, a Cartesian impedance law of the form $\boldsymbol{\tau}_{\text{RL}} = \mathbf{J}^\top \left( \mathbf{K}_p (\mathbf{x}_{ref} - \mathbf{x}) - \mathbf{K}_d \dot{\mathbf{x}} \right),$ with $\mathbf{J}$ the robot Jacobian and $(\mathbf{x}_{ref}, \mathbf{x},\dot{\mathbf{x}})$ the task-space position reference, real position and velocity, equally defines a torque command. Such torque-level parameterizations can be incorporated in the QP without modification. A comparative study of alternative torque-inducing control laws for RL policies is provided in~\cite{rl-torque-laws}.

\subsubsection{TorqueTask}
The TorqueTask enforces the desired joint torques as a linear equality in the QP (Eq.~\eqref{eq:QP}):
\begin{align}\label{eq:QP-torquetask}
\mathbf{A} &= \mathbf{M}(\mathbf{q}), &
\mathbf{b} &= \boldsymbol{\tau}_d - \mathbf{D}(\mathbf{q}, \dot{\mathbf{q}}),
\end{align}
where
\begin{equation}\label{eq:D-hat}
    \mathbf{D}(\mathbf{q}, \dot{\mathbf{q}}) = \mathbf{C}(\mathbf{q}, \dot{\mathbf{q}})\dot{\mathbf{q}} + \mathbf{g}(\mathbf{q}) - \hat{\boldsymbol{\tau}}_{\text{ext}},
\end{equation}
and $\boldsymbol{\tau}_d \in \mathbb{R}^{n}$ is the desired torque (here $\boldsymbol{\tau}_{\text{RL}}$).
This formulation ensures that the QP accelerations $\ddot{\mathbf{q}}^\star$ reproduce the desired torques via inverse dynamics.
 
\subsubsection{Forward Dynamics Task (FDTask)}
An alternative is to convert the desired torque into a reference acceleration using forward dynamics:
\begin{align}\label{eq:QP-fdtask}
\mathbf{A} &= \mathbf{I}, &
\mathbf{b} &= \mathbf{M}(\mathbf{q})^{-1}\bigl(\boldsymbol{\tau}_d - \mathbf{D}(\mathbf{q}, \dot{\mathbf{q}})\bigr),
\end{align}
with $\mathbf{I}$ the identity matrix.
The QP then minimizes the deviation between $\ddot{\mathbf{q}}^\star$ and this reference acceleration.
While straightforward, this formulation requires inversion of $\mathbf{M}(\mathbf{q})$, making it more sensitive to modeling errors in the dynamics, as we include the model into the objective in $\mathbf{b}$.

\section{Experiments}

We evaluate our approach on two different robotic platforms. First, on a Kinova Gen3 7-DoF manipulator, we use an off-the-shelf RL policy trained without safety constraints to demonstrate plug-and-play compatibility with standard RL pipelines. On the Unitree H1 humanoid, we consider a locomotion policy trained with a Safe-RL method, enabling comparison between runtime formal filtering and training-time safety enforcement. All experiments are implemented within the control framework \texttt{mc\_rtc}\footnote{https://jrl.cnrs.fr/mc\_rtc/}, which serves as the primary interface for both robots.

\subsection{Manipulator Experiments}

We use the publicly available PPO policy from~\cite{Le_Lay_Kinova_Gen3_RL_2025}, trained for end-effector reaching on the Kinova Gen3 without safety constraints. Acc-CBF-QP enforces manufacturer-specified joint position, velocity, and torque limits, and self-collision avoidance. The policy runs at 100\,Hz, generating $\mathbf{q}_\text{RL}$ commands. Our controller runs at 1\,kHz to compute $\boldsymbol{\tau}_\text{RL}$, and ensures constraints at the controller frequency. For precise torque tracking and external disturbance estimation, we follow the pipeline described in~\cite{muraccioli2025rss}.

\subsubsection{Nominal conditions} 
\begin{figure}[t]
    \centering
    \includegraphics[width=0.35\textwidth]{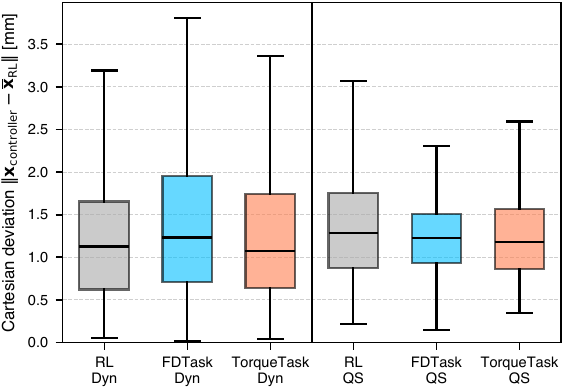}
    \caption{Behavior preservation under nominal conditions. Boxplots of Cartesian deviation relative to the mean RL trajectory, for RL (gray), FDTask (cyan), and TorqueTask (orange) during dynamic (Dyn) and quasi-static (QS) phases. Boxes show the median and interquartile range; whiskers extend to $1.5\times$ IQR.}
    \label{fig:behavior_preservation}
\end{figure}

The robot alternates between two end-effector targets separated by 200\,mm over ten cycles, stopping 5\,s at each target. To quantify RL nominal behavior preservation, we compute the mean Cartesian trajectory of RL over all cycles $\overline{\mathbf{x}}_{RL}$, and evaluate the deviation $\|\mathbf{x}_{controller} - \overline{\mathbf{x}}_{RL}\|$ for RL (intrinsic repeatability), FDTask, and TorqueTask during dynamic and quasi-static phases. RL+QP deviations remain of the same order as intrinsic RL repeatability in both phases (Fig.~\ref{fig:behavior_preservation}). A paired Wilcoxon test on per-cycle RMS deviations detects no significant difference between RL and RL+QP in either phase ($p>0.12$). Non-inferiority tests with a 0.5\,mm margin further confirm that RL+QP deviations are statistically bounded within this tolerance for both tasks and phases, confirming no measurable degradation under nominal, violation-free conditions.

\subsubsection{Out-of-distribution scenario} 
\begin{figure}[t]
    \centering
    \includegraphics[width=0.35\textwidth]{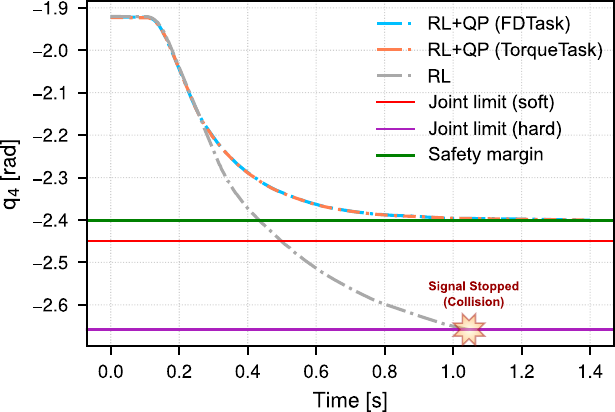}
    \caption{Joint 4 position during OoD execution. RL alone (gray) exceeds the hardware joint limit, triggering a shutdown. Both RL+QP tasks (cyan and orange) strictly respect the joint limit constraints.}   
    \label{fig:comparison_joint_limit}
\end{figure}

Replacing the training initial configuration $\mathbf{q}_0$ with the manufacturer default induces a joint limit violation between joints 3 and 4. The baseline RL policy exceeds joint limits and triggers a hardware safety shutdown, whereas both QP formulations enforce limits and prevent collision (Fig.~\ref{fig:comparison_joint_limit}).

\subsection{Humanoid Experiments}
We evaluate our framework on a more challenging floating-base system, the 19-DoF Unitree H1 humanoid robot. Unlike the manipulation experiments, where the RL policy was trained without safety awareness, here the policy is trained using PPO with the CaT approach~\cite{chane2024cat}. Therefore, this setting provides a comparison between Safe-RL approaches through CaT and our Acc-CBF-QP runtime safety enforcement. This walking policy controls only the leg joints, while the arm and torso joints remain static.
Its observations include the angular velocity and roll–pitch orientation from the IMU, the joint positions and velocities of the legs, the previous actions of the policy, the sine and cosine of a gait phase variable (encoding the walking frequency), and a commanded velocity in $x$, $y$, and yaw. The policy was trained to respect joint position and torque limits, as well as ankle collision avoidance. During the experiments, we kept these limits identical in the QP and added joint velocity limits and self-collisions between the other robot bodies. The policy runs at 40\,Hz, while the controller runs at 400\,Hz. Higher control frequencies than this lead to delays in the Unitree API. 
Unless otherwise specified, all experiments use a null velocity command. Under this command, the policy exhibits drifting locomotion rather than stationary walking. Performance is therefore assessed using the survival duration defined in Section~\ref{subsubsection:h1performance}, rather than velocity tracking accuracy.

\subsubsection{Modeling challenges}
We observed a mismatch of more than 8\,kg between the official URDF file and the real robot mass (over 10\% of the total body weight). We empirically improved the dynamic model by adding known masses and checking position tracking with low PD gains, yielding a closer but not perfect model identification. The same model was used for the training and for the QP-based controller. This imperfect model makes the humanoid a particularly challenging test case for RL+QP: model inaccuracies inevitably degrade the consistency of the dynamics. However, by employing the GMO as a disturbance observer to estimate external forces~\cite{residual-arm}, we mitigate much of the effect of model error.

\subsubsection{Control setup}
The H1 API exposes only a PD interface, precluding direct torque commands as 
used for the Kinova. In RL-only mode, the position reference $\mathbf{q}_{\text{RL}}$ 
produced by the policy is sent directly to the interface. In RL+QP mode, $\boldsymbol{\tau}_\text{QP}$ is converted into an equivalent position command by inverting the PD control law:
\begin{equation}\label{eq:inversePD}
\mathbf{q}_\text{cmd} = \mathbf{q}_{\text{a}}+\mathbf{K}_p^{-1}(\boldsymbol{\tau}_\text{QP}+\mathbf{K}_d \dot{\mathbf{q}}_{\text{a}}),
\end{equation}
which ensures that, when passed through the robot’s PD controller, the resulting motor torques are exactly $\boldsymbol{\tau}_\text{QP}$ (since terms cancel out as in  Eq.~\eqref{eq:tauRL}).

\subsubsection{Torque and acceleration error analysis}
We run each mode (RL only, RL+QP with TorqueTask, and RL+QP with FDTask) for approximately one minute on both the real robot and in simulation, using the \texttt{mc\_mujoco} interface~\cite{mc-mujoco}. 
We measure the torque error $\|\boldsymbol{\tau}_\text{QP} - \boldsymbol{\tau}_\text{RL}\|$ and the forward-dynamics acceleration discrepancy $\|\mathbf{M}^{-1}(\boldsymbol{\tau}_\text{QP} - \boldsymbol{\tau}_\text{RL})\|$, which quantify how strongly the QP modifies the RL command. Under perfect modeling and inactive constraints, TorqueTask should yield zero torque error and FDTask zero acceleration error. Results are summarized in Table~\ref{tab:errors}.

\begin{table}
\centering
\caption{Torque and acceleration tracking errors (mean $\pm$ std [max]).}
\label{tab:errors}
\begin{tabular}{lcc}
\toprule
 \textbf{RL+QP Tasks}& \textbf{$\|\boldsymbol{\tau}_\text{QP} - \boldsymbol{\tau}_\text{RL} \|$ [Nm]} & \textbf{$\|\mathbf{M}^{-1}(\boldsymbol{\tau}_\text{QP} - \boldsymbol{\tau}_\text{RL}) \|$ [rad/s$^2$]} \\
\midrule
\textbf{Torque (HW)} & $2.6 \pm 5.0$ [39.4] & $704.9 \pm 1353.9$ [10381.4] \\
\textbf{Torque (Sim)} & $2.1 \pm 4.2$ [26.7] & $546.1 \pm 1094.7$ [6935.7] \\
\midrule
\textbf{FD (HW)} & $12.4 \pm 24.5$ [273.1] & $698.7 \pm 1414.4$ [14717.7] \\
\textbf{FD (Sim)} & $7.4 \pm 15.2$ [122.9] & $468.2 \pm 994.4$ [6703.3] \\
\bottomrule
\end{tabular}
\end{table}

Large magnitudes reflect the norm of all actuated joints and are dominated by transient spikes during foot-ground impact, where discretized accelerations are large by construction.

The results show systematic differences between the two formulations.
For torque tracking, the TorqueTask reduces the mean error by 79.0\% on hardware and 71.6\% in simulation relative to the FDTask. Tracking statistics are comparable between simulation and hardware, indicating limited sensitivity to modeling discrepancies.
For acceleration tracking, the FDTask achieves lower mean error in simulation (14.3\% reduction relative to TorqueTask), consistent with its objective of minimizing forward-dynamics deviation. However, this advantage disappears on hardware (0.9\% reduction), where model inaccuracies degrade consistency between predicted and realized accelerations.
Overall, the TorqueTask exhibits greater robustness to modeling errors, while the FDTask better preserves acceleration consistency when the dynamics model is accurate. When constraints are inactive, both QP task formulations are identical.

\subsubsection{Computational cost}
On the H1 (19\,DoF, 400\,Hz, on Intel Core Ultra 7 155H), the full per-cycle pipeline averages $0.348\pm0.119$\,ms (TorqueTask) and $0.360\pm0.121$\,ms (FDTask), both under 15\% of the control period.

\subsubsection{Constraint violation rates}
Next, we evaluate constraint satisfaction (Tab.~\ref{tab:constraint_violations}) by measuring the number of constraint violations across all joints during each experiment. 
On the real robot, RL alone yields on average 10.04 violations/s, mostly due to torque limits. RL+QP with TorqueTask reduces this to 0.80 violations/s, while RL+QP with FDTask reduces it to 1.32 violations/s.
In simulation, violations are already rare (3.14 torque limit violations/s for RL alone), but both QP tasks achieve 0 constraint violations. The only remaining violations on hardware are in joint velocities.

\begin{table}
\centering
\caption{Constraint violation rates (violations/s), real robot and (simulation).}
\label{tab:constraint_violations}
\begin{tabular}{lccc}
\toprule
\textbf{Lim.} & \textbf{RL} & \textbf{RL+QP (TorqueTask)} & \textbf{RL+QP (FDTask)} \\
\midrule
$\mathbf{q}$ & 0.02 (0.00) & 0.00 (0.00) & 0.00 (0.00) \\
$\dot{\mathbf{q}}$ & 1.05 (0.00) & 0.80 (0.00) & 1.32 (0.00) \\
$\mathbf{d}_\text{coll}$ & 0.00 (0.00) & 0.00 (0.00) & 0.00 (0.00) \\
$\boldsymbol{\tau}$ & 8.97 (3.14) & 0.00 (0.00) & 0.00 (0.00) \\
\midrule
\textbf{Tot.} & \textbf{10.04 (3.14)} & \textbf{0.80 (0.00)} & \textbf{1.32 (0.00)} \\
\bottomrule
\end{tabular}
\end{table}

\subsubsection{Quantifying Performance and Stability Interference}\label{subsubsection:h1performance}
\begin{figure}[t]
    \centering
    \includegraphics[width=\linewidth]{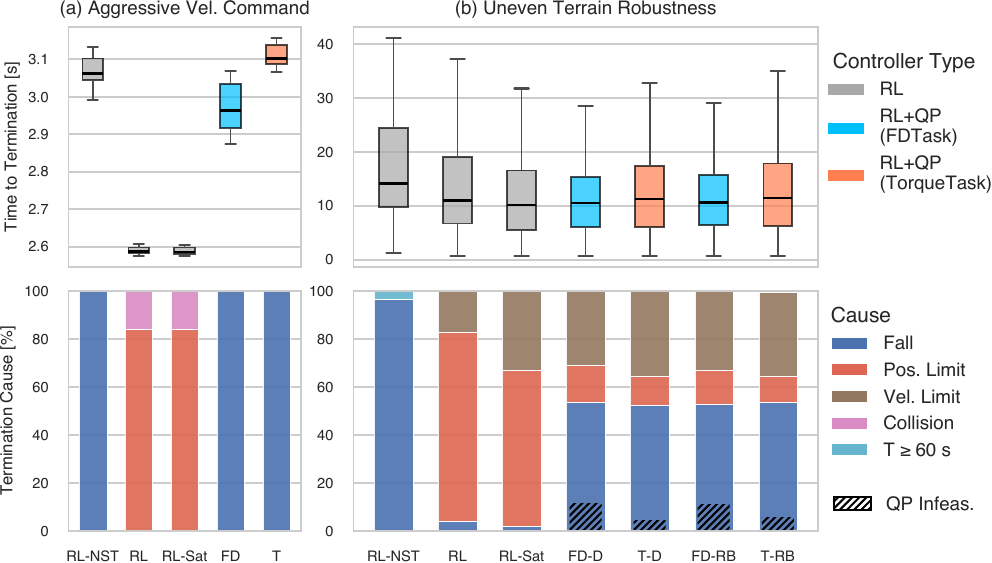}
    \caption{Performance impact of the Acc-CBF-QP across 200 paired simulation trials per controller. \textbf{Top:} survival duration (time to termination). \textbf{Bottom:} termination causes. Left: aggressive velocity command on flat terrain. Right: uneven terrain robustness. Boxplots show median and interquartile range (whiskers: $1.5\times$IQR). Stacked bars report termination categories (fall, joint/velocity limits, collision, timeout); hatched segments indicate QP infeasibility rates for damping (D) and rollback (RB) variants.}
    \label{fig:performance_interference_analysis}
\end{figure}

We evaluate whether embedding the learned locomotion policy within the proposed Acc-CBF-QP degrades performance (Fig.~\ref{fig:performance_interference_analysis}). To evaluate walking performance, we use survival duration as a metric, defined as the time until episode termination. All controllers are tested over 200 \emph{paired} simulation trials with identical randomized initial conditions. Episodes terminate upon: (i) fall (floating-base height below threshold), (ii) 60\,s timeout, or (iii) safety violation (self-collision $e \leq 1\,\mathrm{mm}$, joint violation $\geq1^\circ$, velocity violation $\geq5\%$). Due to hardware risk and sample complexity, all evaluations are conducted in simulation.

\paragraph{Aggressive Velocity Command}

We first consider a flat-ground stress test using a policy trained with reduced ankle-ankle collision margins and commanded with high velocity. This regime stresses feasibility and OoD state rather than disturbance rejection. We compare: RL without safety termination (RL-NST), RL with safety checks, RL with torque saturation (RL-Sat), and the RL+QP variants (FDTask, TorqueTask).
All the variants share the same learned policy and differ only in how constraints are handled. 
RL-NST provides an upper-bound reference by terminating only upon fall or timeout, ignoring safety-related violations. RL activates safety termination conditions without modifying the torque output, isolating the effect of constraint monitoring alone. RL-Sat additionally enforces actuator torque limits via output saturation, representing the hardware-feasible deployment baseline.

RL frequently violates joint limits (84\%) and collision constraints (16\%), yielding a median survival of 2.59\,s. Torque saturation alone does not alter this behavior. Both QP variants eliminate constraint violations entirely and increase median survival to 2.96\,s (FDTask) and 3.10\,s (TorqueTask), corresponding to $\sim$20\% improvement. Paired Wilcoxon tests confirm significant gains over RL-Sat ($p<0.001$). All terminations result from falls, indicating that the QP layer converts unsafe violations into physically admissible failures without degrading task execution.
Notably, RL-NST achieves survival comparable to QP-based controllers while allowing unsafe constraint violations. 
In contrast, the QP layer restores comparable survival while guaranteeing constraint satisfaction, demonstrating that safety can be enforced without sacrificing nominal task performance.

\paragraph{Uneven Terrain Robustness}

We next evaluate the hardware-deployed walking policy on procedurally generated uneven terrain composed of randomly rotated embedded boxes. The policy was trained for rough terrain, but not this configuration. RL-Sat serves as a baseline to ensure identical torque limits.

Unlike previous experiments, QP infeasibility occurred only in this non-planar environment, enabling a comparative evaluation of recovery strategies. Under severe perturbations, QP infeasibility may arise when multiple constraints are mutually conflicting. We evaluate two recovery strategies: (i) switching to damping mode ($\boldsymbol{\tau}_{cmd}=-\mathbf{K}_d\mathbf{\dot{q}}_{\text{a}}$) with saturated torque, until feasibility is restored (FD-D, T-D), and (ii) temporarily reverting to RL-Sat (rollback, FD-RB, T-RB). TorqueTask variants exhibit lower infeasibility rates ($\approx$5\%) than FDTask variants ($\approx$11\%). Rollback marginally increases empirical survival and is the only configuration reaching timeouts, but these differences remain small relative to overall survival times.

Median survival times remain comparable across all controllers (10--11\,s). The raw RL policy predominantly terminates due to joint violations (79\%), whereas QP-based controllers substantially reduce constraint violations and shift failures toward physically plausible falls (42--53\%). 

Across all paired comparisons between FD, Torque, damping (-D), and rollback (-RB) variants, Wilcoxon signed-rank tests detect no statistically significant differences in survival duration in any comparison (all $p>0.08$). Median paired differences remain below 0.35\,s (less than 3\% of nominal survival), with negligible standardized effect sizes ($|d_z|<0.06$). Moreover, 95\% confidence intervals of the paired differences lie entirely within a conservative $\pm2$\,s tolerance band ($\approx 20\%$ of nominal survival), indicating absence of practically meaningful degradation. Thus, the choice of QP formulation or infeasibility handling strategy does not materially affect stability performance on this uneven terrain scenario.

\section{Limitations \& Future Work}\label{sec:limitations}
Despite its practical effectiveness, the proposed framework raises several directions for further investigation.

First, we benchmark only against a CaT-based Safe-RL baseline; broader comparison with methods such as CPO, or integrating Acc-CBF-QP during training, would clarify training-time/runtime complementarity. However, embedding it early may overly restrict exploration, while embedding it late may have limited effect once CaT already suppresses most violations—left for future work.

Second, Sec.~\ref{subsection:formal} bounds safety deviation under bounded disturbance but not closed-loop stability; we observe no instability empirically, but a formal stability characterization of projected RL-controlled systems remains open.

Finally, we handle QP infeasibility pragmatically (damping or torque-saturated rollback), sacrificing formal for statistical guarantees; slack variables, augmented Lagrangian methods, or hierarchical QP prioritizing critical constraints (position, collision) over secondary ones (torque, velocity) are more principled directions, left for future work.

\section{Conclusion}

We introduced Acc-CBF-QP, an open-source runtime safety filter that wraps arbitrary RL policies and enforces position, velocity, torque, and collision constraints at every control cycle, without retraining, via two RL-compatible QP task formulations (TorqueTask, FDTask). Experiments on a 7-DoF manipulator and a 19-DoF humanoid show substantial reductions in constraint violations, no measurable degradation in nominal performance, and longer survival under aggressive commands than a torque-saturated RL baseline. Overall, Acc-CBF-QP provides a practical and scalable mechanism to bridge the flexibility of RL with the strict safety requirements of real-world robotic systems.

\section*{Acknowledgments}
This paper is based on results obtained from the BRIDGE Program (R7-H05), implemented by the Cabinet Office, Government of Japan.

\bibliographystyle{ieeetr}
\bibliography{root}

\end{document}